\documentclass{article}

\usepackage{arxiv}

\usepackage{siunitx}
\usepackage{graphicx}
\usepackage{colortbl}
\usepackage{float}
\usepackage{multirow}
\usepackage{caption}
\usepackage{subcaption}
\usepackage{cite}

\title{Early Prediction of Sepsis using Heart Rate Signals in Wearable Devices with Genetic Optimized LSTM Algorithm}

\author{
  Alireza Rafiei\\
  Department of Mechatronics, School of Intelligent Systems\\
  College of Interdisciplinary Science and Technology\\
  University of Tehran\\
  Tehran, Iran\\
  \texttt{alirezarafiei@ut.ac.ir} \\
  \And
  Farshid Hajati\\
  School of Science and Technology\\
  Faculty of Science, Agriculture, Business and Law\\
  University of New England\\
  Armidale, NSW 2350, Australia\\
  \texttt{fhajati@une.edu.au} \\
  \And
  Alireza Rezaee \\
  Department of Mechatronics, School of Intelligent Systems\\
  College of Interdisciplinary Science and Technology\\
  University of Tehran\\
  Tehran, Iran\\
  \texttt{arrezaee@ut.ac.ir} \\
  \And
  Amirhossien Panahi \\
  Department of Mechatronics, School of Intelligent Systems\\
  College of Interdisciplinary Science and Technology\\
  University of Tehran\\
  Tehran, Iran\\
  \texttt{amirhosein.panahi@ut.ac.ir}
  \And
  Shahadat Uddin \\
  School of Project Management\\
  Faculty of Engineering\\
  The University of Sydney\\
  Sydney, Australia\\
  \texttt{shahadat.uddin@sydney.edu.au}
}

\begin{document}

\maketitle

\begin{abstract}
Sepsis, a critical condition resulting from a dysregulated immune response to infection, significantly impacts mortality, morbidity, and healthcare costs. Effective and timely prediction of sepsis is essential for early intervention, particularly outside the intensive care unit (ICU) where monitoring may not be as rigorous. This study advances the application of machine learning in sepsis prediction by optimizing four established algorithms—LGB, MLP, LSTM, and a novel LSTM-FCN hybrid—for use with wearable devices that monitor heart rate. These models have been specifically enhanced to address the challenges of real-time data processing in non-ICU settings, focusing on improving computational efficiency and reducing memory demands. The architecture of each model was meticulously refined using a genetic algorithm to optimize performance metrics crucial for deployment on wearable technology. Initially developed for a one-hour prediction window, these models were later expanded to four hours using transfer learning techniques. The promising results demonstrate significant potential for wearable devices to support early sepsis detection in diverse clinical environments, highlighting our novel approach to enhancing the practical deployment of machine learning models in healthcare settings.

\end{abstract}

\section{INTRODUCTION}

Wearable technology consists of smart electronic devices worn near the surface of animate skin \cite{duking2018integrated}. They can detect, process, analyze, and transmit information to other devices and data centers. In the last few decades, wearable devices, such as smartwatches and smart bands, have been ubiquitous in modern societies because of their practical applications, especially in health-oriented fields. Such equipment can measure some of the human’s vital signs and ambient features by dint of their high-tech sensors, connecting to other devices and the internet, and analyze information for instant biofeedback to the wearer on an ad hoc basis. As such, healthcare professionals use wearable devices with human activity and sleep pattern recognition ability to acquire behavioral and physiological data to diagnose, manage, and treat acute and chronic diseases \cite{chiauzzi2015patient, mercer2016acceptance}. Studies have shown that remote healthcare monitoring and early warning can potentially improve the quality of health services and reduce patient inconvenience \cite{appelboom2014smart, darwish2011wearable}.
Practical implementation of Internet of Things (IoT)-based telemedicine is burgeoning to prevent disease, predict difficulties at early stages, and generally promote global health \cite{albahri2021iot, rafiei2023meta}. Leveraging the power of Artificial Intelligence (AI) in telehealth not only aids physicians in making data-driven and real-time decisions as a critical factor in providing better health outcomes, but it also can produce primarily warn to individuals. With the accelerated increase of wearable gadget technologies and unprecedented algorithms, individuals rein their well-being, which assists them and physicians in diagnostics and analysis of ailing in primary steps.
Sepsis is introduced as a potentially life-threatening and highly complex condition that leads to multi-organ dysfunction and exhibits numerous morbidity and mortality rates \cite{singer2016third}. Sepsis occurs when the response of the host body’s immune and inflammatory system injures tissues and internal organs instead of fighting infections \cite{shankar2016developing}. Due to early associated symptoms with sepsis, which may be familiar with numerous other clinical conditions, accurate diagnosis of sepsis in the early stages is an uphill struggle for even skilled physicians \cite{jones2010lactate}. According to the World Health Organization’s (WHO) first global report on sepsis, 49 million individuals were estimated to be affected by sepsis, which causes approximately 11 million potentially avoidable worldwide deaths each year \cite{world2020global}. Centers for Disease Control and Prevention (CDC) reported that sepsis develops outside the hospital for almost 80 percent of patients \cite{centerssaving}. Overall, the epidemiology of sepsis in out-of-hospital patients is complex and varies widely across different populations and healthcare settings. A study conducted in Korea found an incidence rate of 28.7 cases per 1000 people each year \cite{kim2019epidemiology}. According to research carried out in the United States, sepsis accounts for a significant portion of emergency department visits and hospitalizations among adult patients \cite{rhee2017incidence}. The study also revealed that patients who received prehospital medical care had lower odds of dying in the hospital. Sepsis survivors encounter severe long-term problems in the form of growth post-discharge mortality, cognitive and physical impairment, and mental disorders \cite{iwashyna2010long}. This tremendous mortality rate is often related to late diagnosis, inadequate infection prevention, and improper clinical management, which can influence vulnerable populations, such as pregnant and older adults \cite{world2020global}.

Sepsis care is expected to be most consequential in the earliest phase of treatment, so the aim is to diagnose sepsis on admission and receive initial care in the emergency department. A one-hour delay in treating sepsis has a consequence of a 4\% to 8\% increment in mortality probability. Hence, early sepsis-onset prediction and required antibiotic intervention are the main steps for ameliorating sepsis outcomes \cite{kumar2006duration, seymour2017time}. To meet this challenge, physicians introduce various diagnosis criteria and scoring systems \cite{jaffer2010cytokines, ms1995natural, subbe2001validation}. The majority of criteria for diagnosing and treating sepsis have been developed for patients in hospital and intensive care unit (ICU) settings. However, it is important to note that up to 50\% of sepsis patients initially come into contact with healthcare providers in the pre-hospital setting \cite{seymour2017time}. The newest sepsis definition, Sepsis-3, was introduced in 2016, and proposed the quick Sequential Organ Failure Assessment (qSOFA) method for rapidly identifying adult patients suspected of infection in out-of-hospital and emergency departments, as well as the SOFA scoring system, which is used chiefly for tracking patients’ status in the ICUs \cite{singer2016third}. The qSOFA utilizes three district criteria, assigning one point for systolic blood pressure (SBP $\leq$ 100 mmHg), high respiratory rate ($\geq$ 22 breaths per min), or altered mentation (Glasgow coma scale < 15). Although \cite{singer2016third} promulgated that qSOFA is less robust than SOFA score due to no requirement of laboratory tests, it can be evaluated rapidly and repeatedly.

The abundance of wearable devices capable of accurately measuring heart rates could be synergetic for transmission of early sepsis onset prediction obligation out of wards. In this paper, we employed four distinct machine learning (ML) approaches for real-time sepsis onset prediction using just low-frequent heart rate signals to install as an auxiliary algorithm on wearable devices. For this reason, an LSTM optimized network with a genetic algorithm (GA), which is trained using the ICU’s Electronic Health Records (EHRs), has shown superior performance. The genetics algorithm is used for the optimization of architectural parameters based on three diverse factors: model performance, the complexity of calculations, and primary storage. More aspects of our research are presented in the succeeding sections: Section 2 investigates the state-of-the-art intelligence system for sepsis prediction; Section 3 deliberates the proposed approach to figure out an appropriate model to predict sepsis onset out of ICU and ward using available features; Section 4 presents the results which are discussed in details in Section 5.

\section{RELATED WORKS}

ML and traditional methods demonstrate significant performance in helping physicians recognize acute clinical situations \cite{panahi2021fcod, rajkomar2019machine}. Related work on pattern recognition, control, and medical AI provides additional context for our modelling choices \cite{
AbdoliHajati2014,
Ayatollahi2015,
BarolliAINA2024,
BarolliBWCCA2019,
BarolliWAINA2019,
Barzamini2012,
CremersACCV2014,
Fiorini2019,
Hajati2017Surface,
Hajati2006FaceLocalization,
Hajati2010PoseInvariant,
Hajati2017DynamicTexture,
Mahajan2024,
Pakazad2006FaceDetection,
Sadeghi2024COVID,
Shojaiee2014Palmprint,
Sopo2021DeFungi,
Tavakolian2022FastCOVID,
Tavakolian2023Readmission,
Wang2022SoftwareImpacts,
KarimiRezaee2017Helmholtz,
MohamadzadeRezaee2017Antenna,
Ramezani2024Drones,
Rezaee2008GeneticSymbiosis,
Rezaee2010FIR,
Rezaee2017PID,
Rezaee2017Penetrometer,
Rezaee2017MPC,
RezaeeGolpayegani2012,
RezaeePajohesh2016,
Sadeghi2024ECG,
Taghvaee2014Metamaterial,
Tavakolian2022SoftwareImpacts,
Gavagsaz2018LoadBalancing,
Rezaee2014FuzzyCloud,
Sarvghad2011ThinkingStyles,
Shahramian2013Leptin}. One of the primary and influential research in this field was reported in 2016 when \cite{desautels2016prediction} introduced the InSight algorithm to predict sepsis onset using vital signs and age. This ML-base model was designed for sepsis onset prediction in ICUs until four hours in advance and could achieve an AUROC of 0.57 for this aim. Two years later, \cite{mao2018multicentre} validated and improved the InSight algorithm using the transfer learning technique and a new dataset for predicting sepsis onset based on three different gold standards. While they used six distinct vital signs as the input, they were able to ameliorate the AUROC InSight to 0.85. Meanwhile, \cite{nemati2018interpretable} proposed the AISE Algorithm, which can predict the onset of sepsis 12 hours in advance. This model used low-frequency EHR data and high-frequency heart rate to accomplish its task, attained an AUROC of 0.84. Furthermore, \cite{thakur2018neonatal} applied binary logistic regression to improve and analyze two invasive and non-invasive prediction networks based on the MIMIC III dataset. For this aim, they developed an Android application to estimate the risk of sepsis onset. Their results illustrated that the non-invasive parameters are more efficient than the invasive parameters, gaining an AUROC of 0.77 and 0.82 for the latter and the former networks, respectively. Besides, as the unusual body temperature is one of the compelling features of sepsis prediction, \cite{al2018early} suggested the usage of thermography as a mechanism for diagnosing sepsis. This idea was developed by evaluating patterns of continuous body temperature with a fully automated unsupervised ML structure for discrimination between outer temperatures of the ear and its core. By implementing a fuzzy C-Means clustering technique, they classified the temperature variation into distinct groups.
Over the next two years, \cite{lauritsen2020early} designed an LSTM-CNN model for early sepsis prediction and compared its efficiency with gradient boosting and multilayer perceptron networks. They utilized a specific Danish dataset containing unstructured data elements and medical imaging as well as vital signs and laboratory tests. This dataset comprised both ICU and ward patients and was labeled with SIRS criteria \cite{levy20032001}. With the proposed network, they attained an AUROC of 0.79 for a ten-hour prediction window. Additionally, \cite{kok2020automated} developed a temporal convolutional network to predict sepsis using the PhysioNet 2019 Challenge dataset. By applying per-patient evaluation metrics, they recorded an AUROC of 0.91. Besides, they evaluated their model using three different validation methods to confirm its robustness.
In 2021, \cite{rafiei2021ssp} proposed the SSP model to predict sepsis onset up to 12 hours in advance for patients admitted to ICUs. They used vital signs, demographic values, and laboratory tests in two separate modes and applied PhysioNet 2019 Challenge dataset for developing their model. The performance of the SSP model achieved 0.92 AUROC for a four-hour prediction window. Moreover, \cite{shashikumar2021deepaise} presented the DeepAISE algorithm with the aim of practical implementation. They used several distinguished datasets with the gold standard of sepsis-3. The performance of this model, which was tested in three distinct healthcare systems, shows AUROC of between 0.87 and 0.90 using six vital signs in a four-hour sepsis onset prediction window for ICU patients and a relatively low false alarm rate. Recently, \cite{liu2022machine} introduced a real-time sepsis prediction framework to address the shortcomings in dynamic changing adaptability and the reliance on a pre-set threshold. In this regard, they developed a machine learning-based model, including a random forest and a multilayer perceptron (MLP) architecture, and integrated it into a partially observable Markov decision process model. While they provided a test set with the same ratio of sepsis and non-sepsis ICU patients, the proposed framework, which used a random forest method, achieved an AUROC of 0.76.
In another work, Smith et al. (2021), who utilized a variety of machine learning models to predict sepsis onset from clinical data, demonstrating substantial improvements in prediction accuracy (https://doi.org/10.1002/int.22370). Their findings underscore the potential of advanced analytics in enhancing patient outcomes in critical care settings. Similarly, Johnson and Lee (2022) explored the integration of artificial intelligence with electronic health records to predict sepsis, revealing that real-time data processing could play a crucial role in early sepsis detection (https://doi.org/10.1016/j.dcan.2022.06.019). These studies provide a solid foundation for our research, which builds upon these methodologies by optimizing machine learning algorithms for use in non-ICU environments, particularly focusing on the unique challenges posed by wearable device data.
To date, studies are primarily focused on in-hospital, especially ICU patients, and applied multifold features to predict sepsis onset. Thus, there is a lack of ML approaches that are optimized for early sepsis prediction with continuously measurable features out-of-hospital. The challenging issues of out of medical-related environments involve the shortcomings in accurately measured features, the restricted storage over a period of time, and the limited power for computing hardware. Therefore, the potential prospective approaches should be able to work with as limited input features as possible and be memory-efficient for real-world implementation.

\section{MATERIAL AND METHODS}
\subsection{DATABASE}

We have used the PhysioNet/Computing in Cardiology Challenge 2019 dataset to develop our models since it is one of the most comprehensive and widely used publicly available sepsis prediction datasets. It comprises EHRs of 40,336 patients who were admitted to the ICU, of which 20,336 records are related to the Beth Israel Deaconess Medical Center (BIDMC) and 20,000 records related to the Emory University Hospital (EUH). Each EHR contains eight vital signs and 26 results of the laboratory tests on an hourly and daily basis, as well as six demographic details of ICU patients. Nevertheless, due to the sparsity affected by the lost values, there is no particular frequency in capturing the dataset. Healthcare experts had reported this information during the previous decade when a patient was admitted to the ICU, including the approval from the relevant Institutional Review Boards. All features of the data were consolidated into hourly intervals; for instance, multiple heart rate readings within an hourly timeframe were summarized as the median heart rate. Also, the EHRs just contain patients who developed sepsis after being admitted to ICUs \cite{reyna2020early}. The dataset provides an hourly label of “sepsis” or “non-sepsis” for each patient during their length of stay using the Sepsis-3 gold standard. The onset time of sepsis with two or more point variations is presumed infection in the SOFA score from 24 hours to 12 hours after the mistrust time. Also, clinical mistrust of infection was classified as the first timestamp of antibiotic treatment  \cite{shankar2016developing,singer2016third}.

\subsection{PRE-PROCESSING}

The objective of this study is to predict sepsis in settings outside of ICUs and wards, specifically considering the application of models that could be utilized with data from wearable devices in future studies. However, it's important to note that for this study, we utilized data sourced from standard clinical monitoring equipment as provided by the MIMIC dataset, which records heart rate among other clinical parameters. This dataset includes data typically collected in ICU settings, which is not directly from wearable devices but shares similar characteristics in terms of single-feature focus and frequency of data recording. In our preprocessing flow, we initially discarded all features from the electronic health records (EHRs) except for heart rate due to the focus on this single metric which is commonly available and accurately measured even in less complex devices. The EHRs were then subjected to a plausibility filter to remove readings with heart rates below 15 or above 300 beats per minute, considered anomalous. For our analysis, we included only patients over the age of 14 who were admitted to the ICU for more than 12 hours. For patients labeled as "non-sepsis," we extracted the last 12 hours of their records, irrespective of the total length of stay. Conversely, for those meeting sepsis criteria, we extracted a twelve-hour record centered around one and four hours before the onset of sepsis, respectively. These records were then appropriately labeled as "sepsis" or "non-sepsis" based on the clinical data provided. This approach allows us to refine our predictive models in a controlled setting, using data that, while not from wearable devices, approximates the type of single-feature data that wearable devices would typically collect. This clarification ensures the transparency of our methodology and the context in which our findings can be applied to potential future applications involving wearable technology.

\begin{figure}[!h]
\centering
\captionsetup{justification=centering}
\includegraphics[width=6cm]{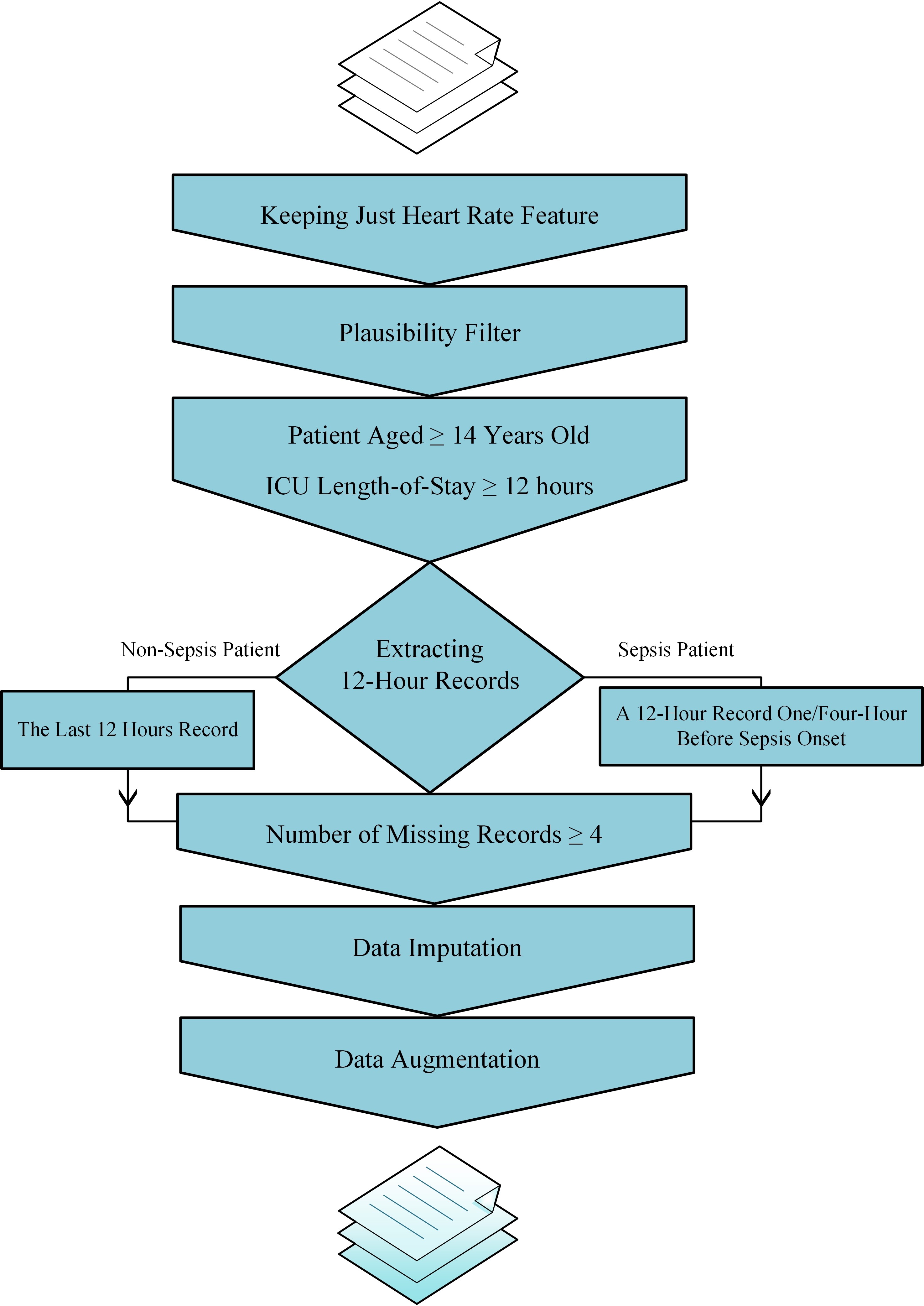}
\caption{The pre-processing workflow.}
\label{fig 1}
\end{figure}

There were missing values for some of the hourly records. More specifically, 9.9\% of the heart rates were missed in the dataset. To tackle this problem, first, we eliminated the processed EHRs in which more than 4 hours of records were unavailable. Second, we applied the forward-fill imputation method to the rest of the EHRs that carried forward the most recent record to fill in the missing value.
Roughly 6\% of the pre-processed twelve-hour heart rate EHRs had the label of sepsis. Therefore, we utilized the noise injection data augmentation method to balance the dataset \cite{wen2020time}. Noise injection was implemented randomly on one to eight heat rate records of the resulting EHRs. The added noise had a number in the interval of $±$4 beats with a Gaussian distribution. We continued the data augmentation method until 30\% of the dataset involved the “sepsis” label. Figure \ref{fig 1} represents the pre-processing workflow of this study.

\subsection{PROPOSED MODELS}

To attain the objective of implementing an appropriate deep-learning model on wearable devices, we should consider several contributing factors. Given that these devices have limited memory, power, and batteries, the model must operate as efficiently as possible. Of note, the proposed model should be able to predict the onset of sepsis with the minimum number of measurements, restricted calculations, and memory usage, robust in the face of noises, and reliable performance. We have developed four district ML approaches to predict sepsis using the heart rate feature considering their efficiency. Moreover, we have utilized a genetic algorithm in our proposed methods to help discover the most optimized structures. The critical advantage of this approach is the GA-based solution’s capacity to find an answer for highly complex nonlinear problems. It is well-established that some neural networks may be trapped in local minima and have a slow learning rate \cite{goodfellow2016deep}. Due to GA’s multidirectional searching system, the neural network can improve its approximation capability to be converted to a global minimum. To select fine-tuned parameters, we have considered a specific criterion for all ML models: in every attempt, model performance, size of the model, and the complexity of calculations (execution time) are regarded as determinating factors. Eventually, these factors were ranked, and the best average rating was chosen to build up the models. Primarily, these models were developed for the sepsis onset prediction one-hour in advance. After that, they have been generalized to a four-hour prediction window by employing the transfer learning method.

\subsubsection{Light gradient boosting (LGB)}

LGB is a distributed, high-performance, and fast framework based on a decision tree algorithm used for vast ML applications \cite{ke2017lightgbm}. In the first baseline, we have constructed an LGB model by considering a one-hour prediction window EHRs involving the heart rate of patients as the defining feature. To find the optimized parameters of the LGB model (i.e., number of leaves, maximum bin, and learning rate), we have utilized the GA algorithm. In every attempt, we trained the model with GA-proposed parameters and considered its determinating factors in the face of nailed validation data as feedback. The outcome of each run was used to rate that attempt using the aforementioned criteria.

\subsubsection{Multilayer perceptron (MLP)}

Contracting an MLP as a standard class of multilayer feedforward neural network was the second experience. MLP has the ability to learn the training data representation to decipher the best relation to the desired prediction of the output variable \cite{ramchoun2016multilayer}. We have implemented a deep approach to MLP with four layers in this study. At the first step, one-hour prediction window data was utilized for training the MLP model, and the number of neurons in each layer was optimized using the GA algorithm, with the same strategy as the LGB experiment. To be specific, 100, 148, and 74 neurons for the first, second, and third hidden layers have been probed by the GA algorithm as the optimized number to fulfill our criteria. For the last layer, we have considered a solitary neuron to estimate the probability of sepsis onset. The ReLU activation function \cite{agarap2018deep} was applied for all three hidden layers. This model has been trained using the mean square error loss function and Adam optimizer with mini-baches of 32, while the training data has been shuffled every 390 epochs.
Afterward, we used the one-hour prediction window network to develop a four-hour predictive model. That is, we have employed a transfer-learning technique to implement an MLP model to predict sepsis utilizing the heart rate feature four hours in advance. Using the previously obtained setting, the model reached convergence in the face of new data during 50 epochs.

\subsubsection{Long short-term memory (LSTM)}

\begin{figure}[!h]
\centering
\includegraphics [width=14cm , height = 9cm] {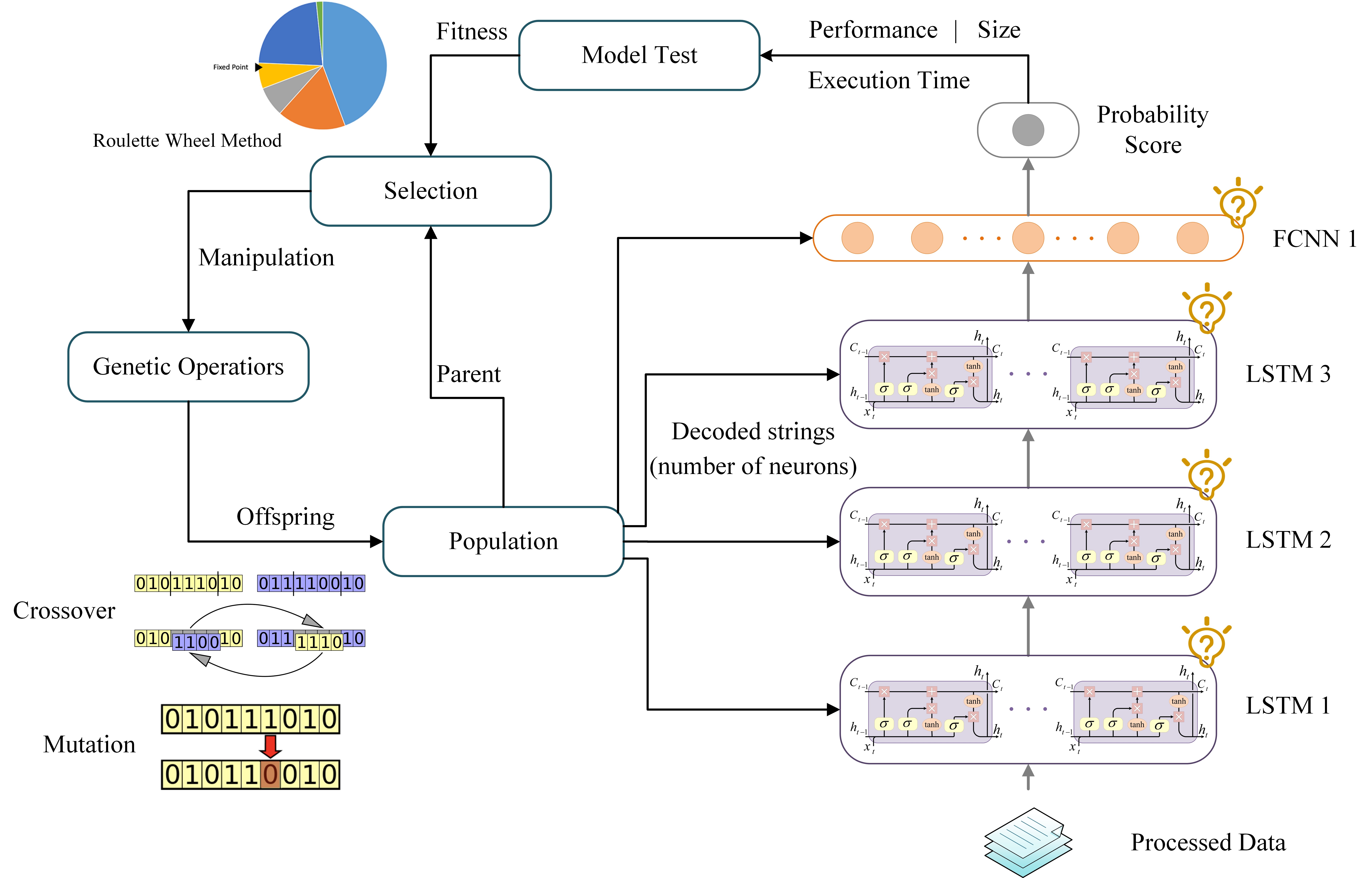}
\captionsetup{justification=centering}
\caption{The workflow of finding the optimal number of neurons for the LSTM model.}
\label{fig 2}
\end{figure}

LSTM models have a large number of effective parameters, more importantly, the number of neurons and layers that need to be adjusted and modified \cite{hochreiter1997long}. Nevertheless, computational time and cost restrictions make it difficult to tune the parameters and determine the optimal scheme of its structure. Therefore, discovering these parameters of LSTM networks that notably alter the performance of models is considerable. The proposed model of this study employs a GA-based optimization to explore the optimal number of neurons for layers of deep learning architecture. As the model’s size and the complexity of its calculations are also influential in addition to the model’s performance for the aim of implementation on wearable devices, the GA has been applied to manage the optimal structure of hidden layers of the LSTM model for fulfilling the desired criteria at the same time.
The proposed LTSM model’s structure consists of three LSTM layers followed by a fully-connected layer. At the end of the model, a solitary neuron estimates the probability of sepsis onset. When the GA suggested the number of neurons for each layer, the model started training in twenty epochs. To be sure about convergence, if the average of the first two epoch loss values of train data was less than the last two, we discarded GA’s proposal. The fitness function of the GA has been calculated using the mentioned criteria. In this regard, the accuracy and specificity of the trained models with proposed parameters on a validation set have been utilized for analyzing the models’ performance. In addition, the execution time of a prediction has been considered to represent the complexity of its calculations, and the primary usage criteria have been based on the occupied size of a model. Of note, the execution time was determined by averaging the prediction for every EHR in the validation set. We have created eight variants for each four required to optimize layers to discover the best number of neurons in the range of 0-255 and then squeeze them together to form a gene. Furthermore, the variable of the GA was initialized as Bernoulli random variables, and a roulette wheel selection algorithm was applied as the selection method. Also, the crossover and mutation operations were used in the optimization workflow. The GA proposal has been evaluated in the last step and transferred to the selection pool, where the cycle begins anew. Figure \ref{fig 2} depicts the model’s architecture and the GA workflow to probe the most optimized number of neurons. After implementing the GA, 48, 108, 52, and 20 neurons were selected for the three LSTM and one fully connected layer, respectively. While the activation function of all layers was chosen as tanh \cite{lau2018review}, the Adam optimizer was considered for the model. Having used a batch size of 32, the optimal model was trained in 190 epochs. The GA algorithm was implemented using the DEAP python library, and the deep model implantation was on Keras 2.3.1 with a TensorFlow 2.2.0 backend using Google Colaboratory.

We have again trained this model to fine-tune its weights for a four-hour prediction window using the transfer-learning technique. Hence, we have passed the pre-processed four-hour prediction window records from the one-hour prediction window deep model and trained the model for 50 epochs.

\begin{figure}[!hth]
\centering
\includegraphics [width=10cm , height = 6cm] {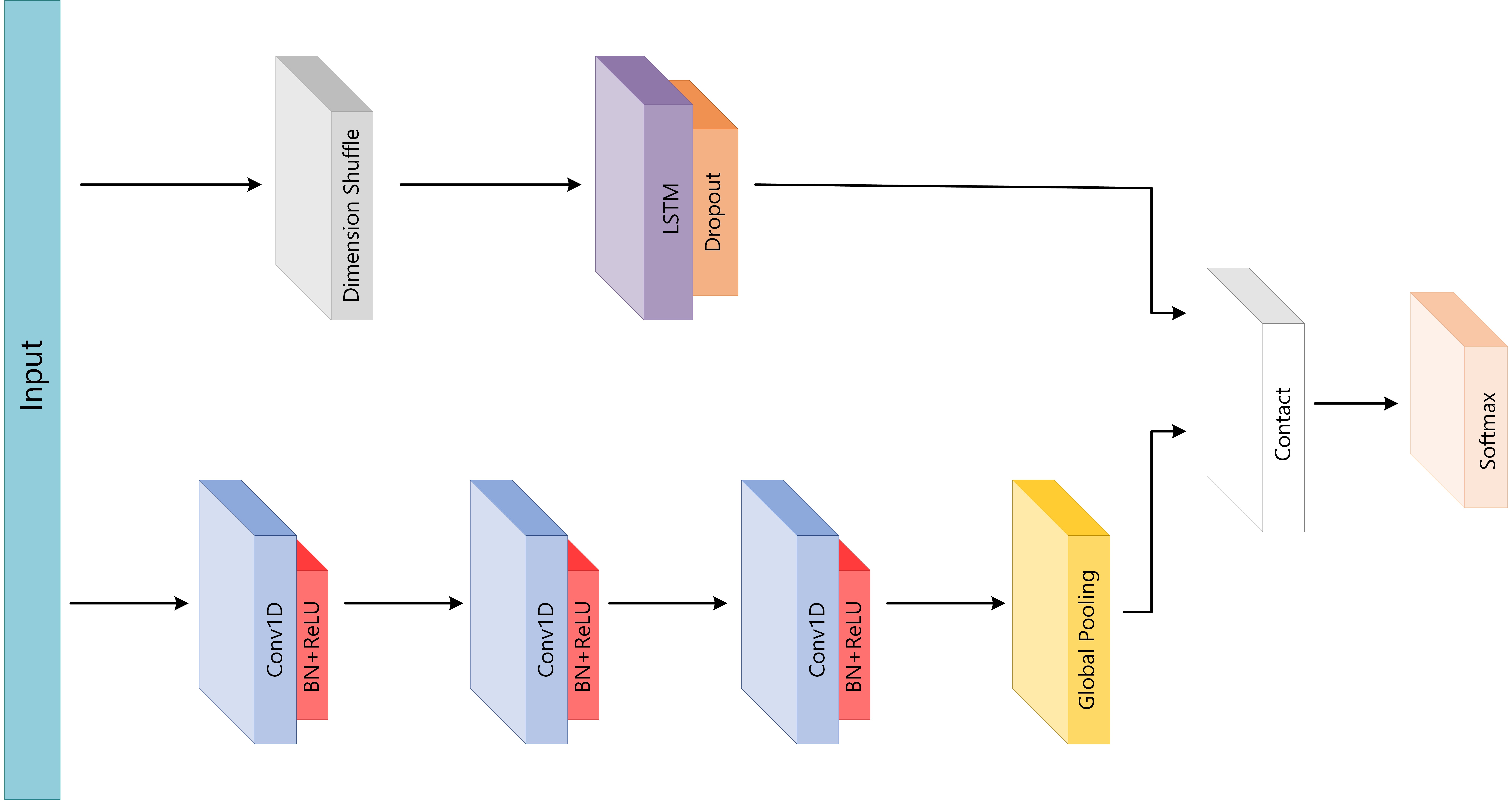}
\captionsetup{justification=centering}
\caption{The architecture of LSTM-FCN model.}
\label{fig 3}
\end{figure}

\subsubsection{LSTM-FCN}
LSTM Fully Convolutional Network (LSTM-FCN) can achieve state-of-the-art performance on the classification of time series sequences \cite{karim2017lstm}. This model does not require time-consuming, heavy processing, or feature engineering on the dataset. LSTM-FCNs can augment FCN models with nominal parameter enhancement to increase their performance appreciably. In this experiment, we have used a customized LSTM-FCN model to predict sepsis onset utilizing a time series of heart rates. For this aim, we have preserved the proposed structure of the LSTM-FCN model \cite{karim2017lstm}, but its parameters were customized to reach the highest performance of sepsis prediction (Figure \ref{fig 3}). So, we have considered 32 neurons for the LSTM layer followed by a 40\% dropout layer, which gives input from a dimension shuffle layer. Three parallel temporal convolutional blocks have proposed filter sizes of 16, 32, 16, and strides of 3, 3, and 2, respectively. Each block comprised a temporal convolutional layer, accompanied by batch normalization with a momentum of 0.99 and epsilon of 0.001, and used a ReLU activation function. After convolutional blocks, global average pooling was served. Eventually, the output of the global pooling layer and the LSTM block were concatenated and conveyed onto a softmax classification layer to give an early prediction. It should be noted that the GA algorithm was not employed in this approach.

The developed LSTM-FCN was trained with the same hyperparameters and software configuration of the proposed LSTM model and categorical cross-entropy loss function in 300 epochs. It first exposed a one-hour prediction window, and then, using the transfer-learning method, the weights of the model tuned in 100 epochs for a 4-hour prediction window.

\section{RESULTS}
\subsection{LGB}

We have analyzed and compared the performance metrics and specifications of the developed models (Table \ref{table 1}). Figure \ref{fig 4} illustrates the ROC curve, precision-recall (PR) curve, and calibration curve for the one-hour prediction window of the models on the test dataset, while Figure \ref{fig 5} demonstrates the AUROC and average precision (AP) of one- and four-hour prediction windows. The LGB model achieved an AUROC of 0.825 and a mAP of 0.637 when evaluated for a one-hour prediction window. These figures for a four-hour prediction window were 0.804 and 0.610, respectively. Focusing on the calibration curve (Figure 5c), we can see that the LGB’s prediction of sepsis development risk is always higher than the observed frequency of the onset of sepsis. In addition, the architecture size of the LGB model is 196 kilobytes (kb), and it can do the needed calculations in 0.004 milliseconds (ms).

\subsection{MLP}

In evaluating the MLP model, AUROC of 0.686 and 0.623, and AP 0.497 and 0.414 for one- and four-hour prediction windows were achieved. As clearly can be seen in Figures \ref{fig 4} and \ref{fig 5}, the MLP model had the lowest performance in both windows in comparison with all other methods, while it had the second-highest architecture size. However, it possesses the lowest execution time (0.09 ms) of all deep learning models.

\subsection{LSTM}

\begin{figure}[!ht]
\captionsetup{justification=centering}
     \centering
     \begin{subfigure}[b]{0.44\textwidth}
         \centering
         \captionsetup{justification=centering}
         \includegraphics[width=\textwidth]{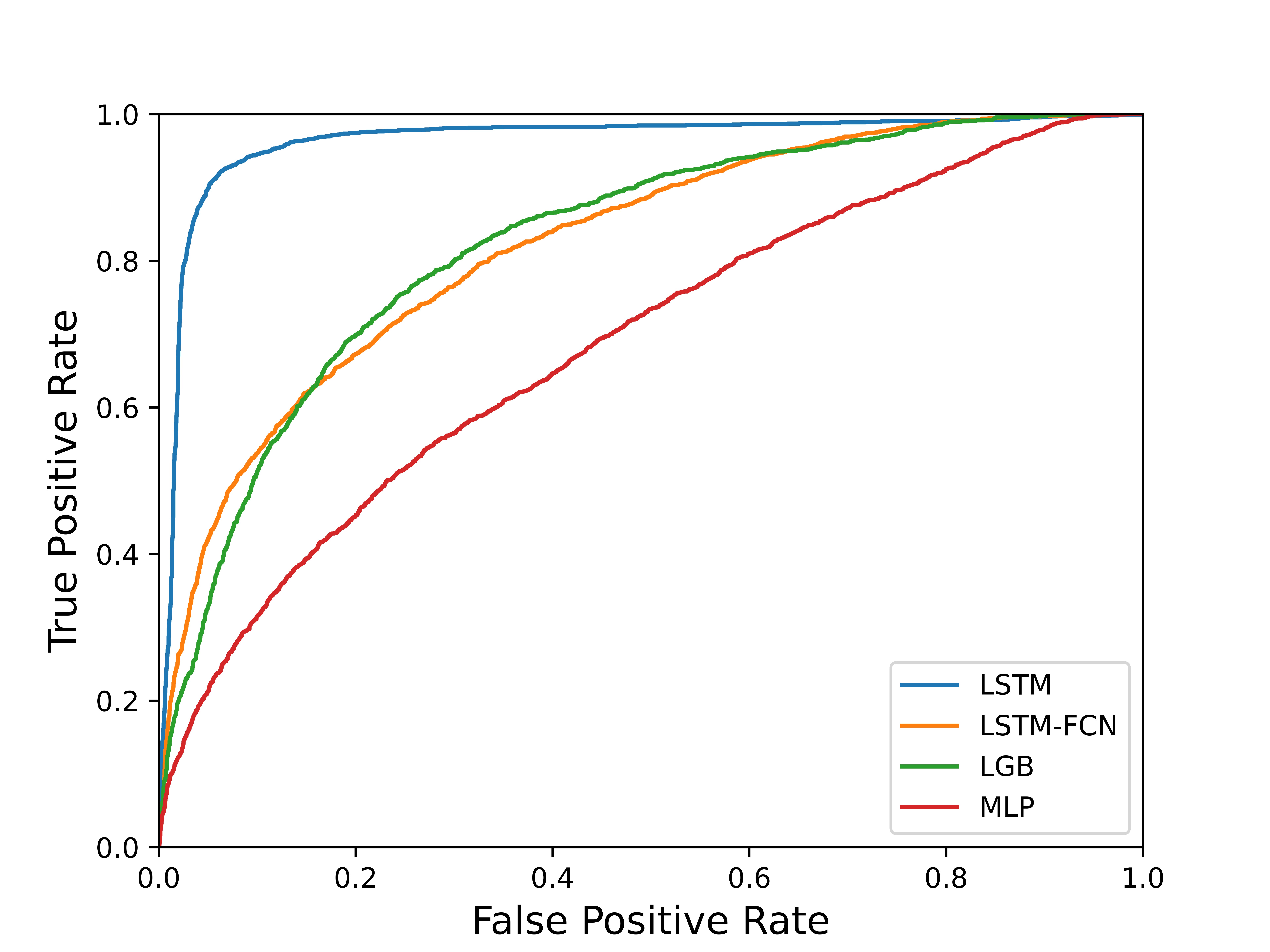}
         \caption{}
         \label{5a}
     \end{subfigure}
     \hfill
     \begin{subfigure}[b]{0.44\textwidth}
         \centering
         \captionsetup{justification=centering}
         \includegraphics[width=\textwidth]{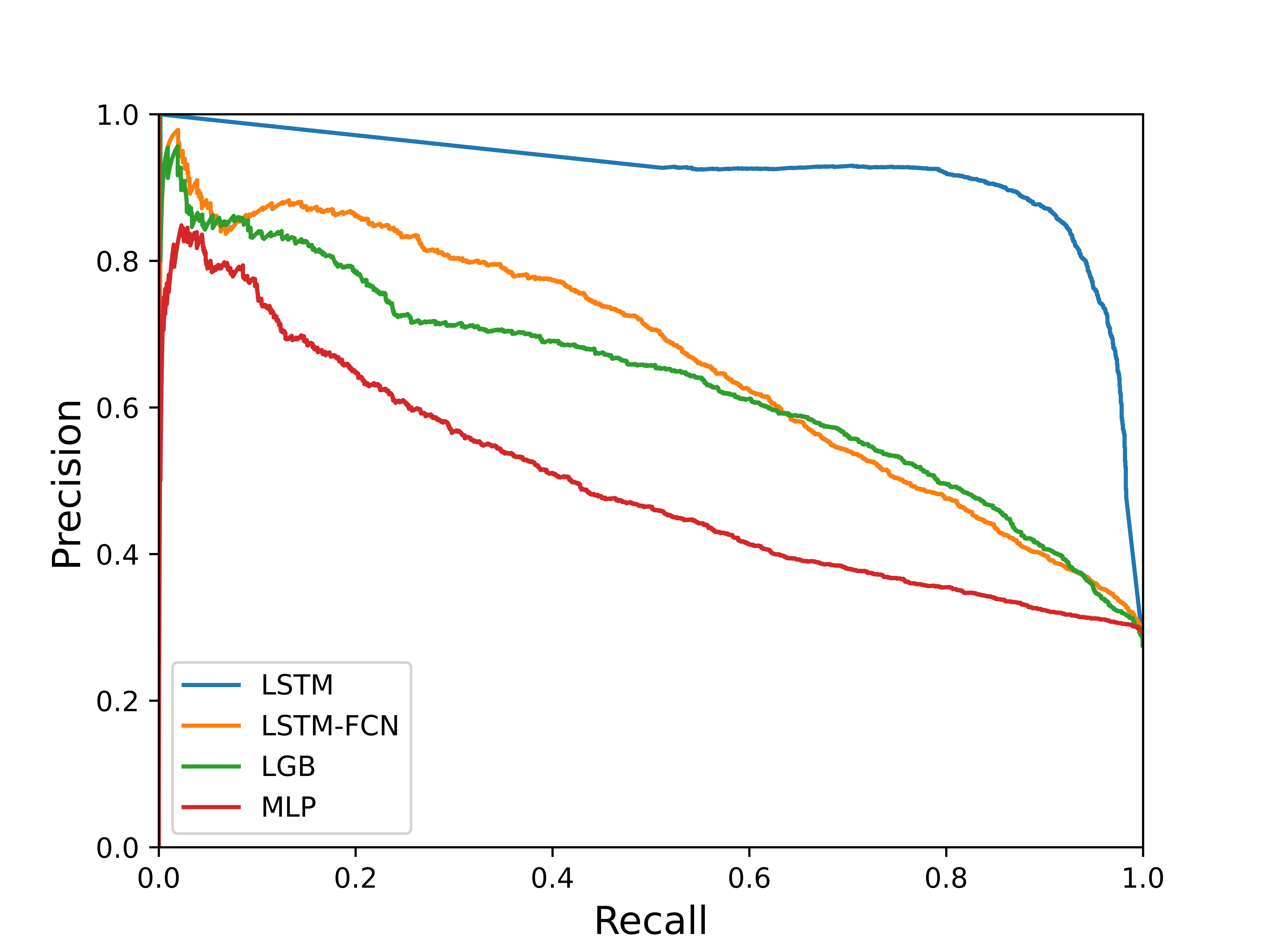}
         \caption{}
         \label{5b}
     \end{subfigure}
     \hfill
     \begin{subfigure}[b]{0.44\textwidth}
         \centering
         \captionsetup{justification=centering}
         \includegraphics[width=\textwidth]{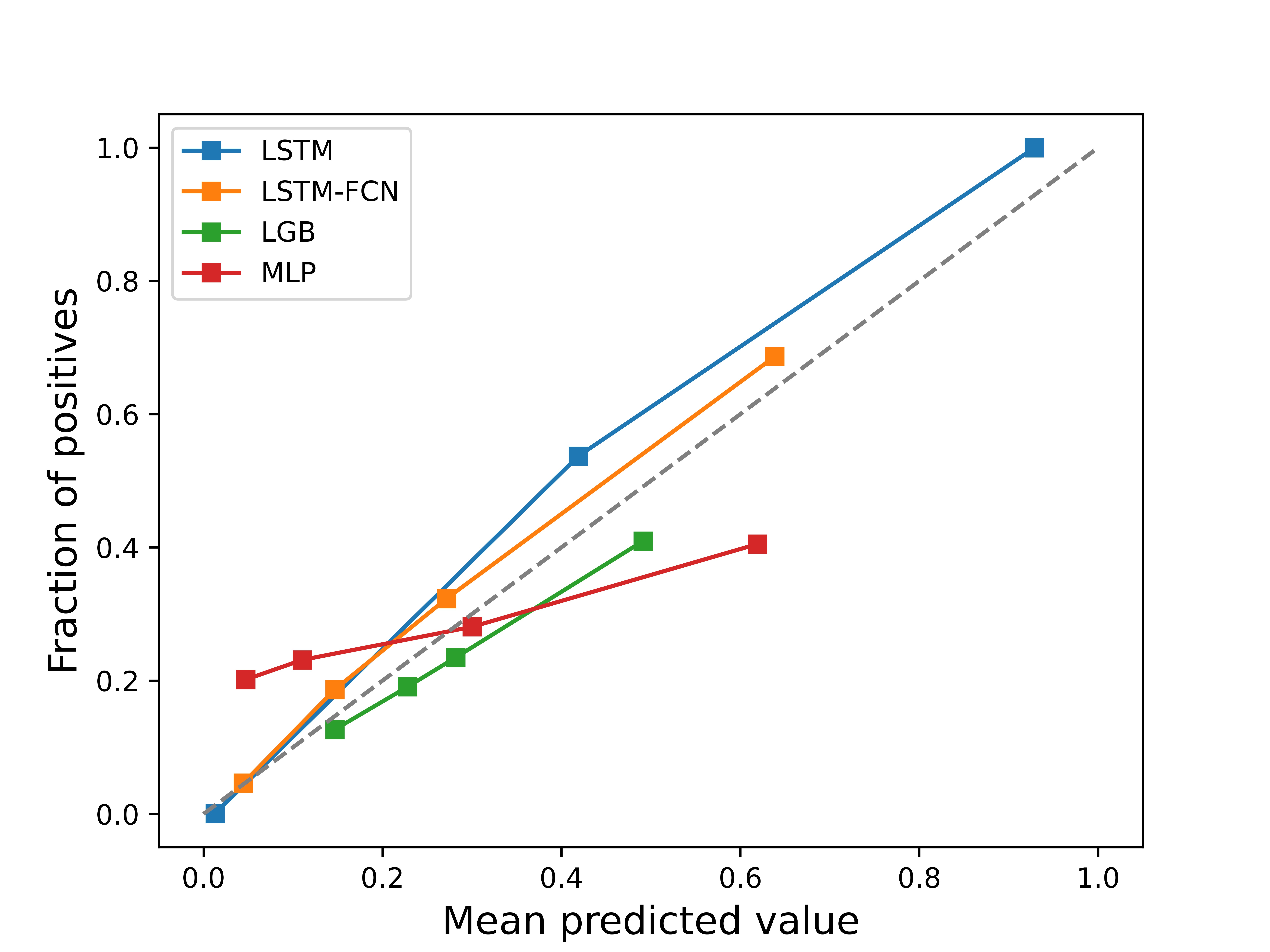}
         \caption{}
         \label{5c}
     \end{subfigure}

        \caption{Results from the test dataset were evaluated one hour before sepsis onset: (a) ROC curves; (b) PR curves; (c) Calibration curve.}
        \label{fig 4}
\end{figure}

\begin{figure}[!ht]
\captionsetup{justification=centering}
     \centering
     \begin{subfigure}[b]{0.44\textwidth}
         \centering
         \captionsetup{justification=centering}
         \includegraphics[width=\textwidth]{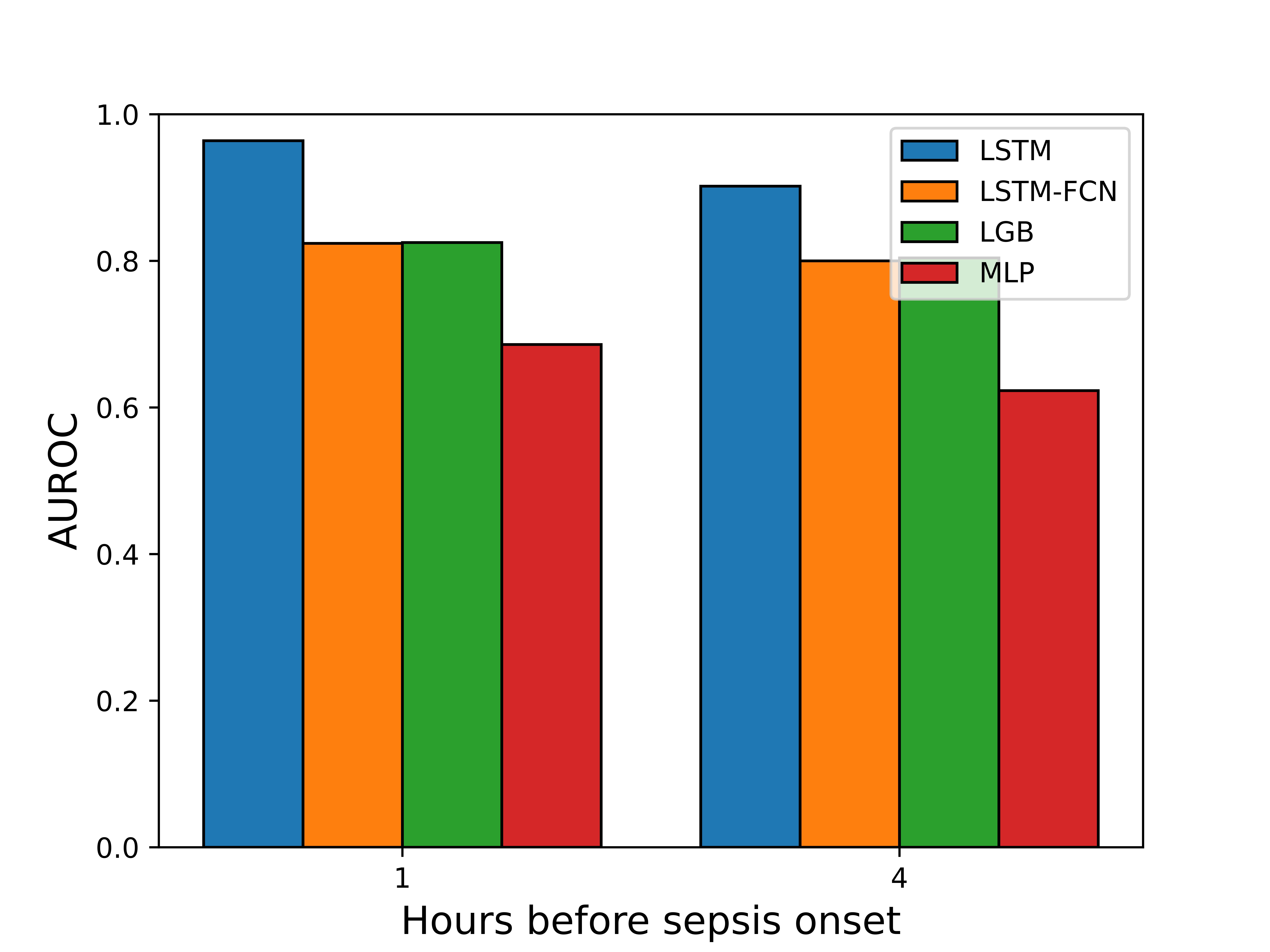}
         \caption{}
         \label{6a}
     \end{subfigure}
     \hfill
     \begin{subfigure}[b]{0.44\textwidth}
         \centering
         \captionsetup{justification=centering}
         \includegraphics[width=\textwidth]{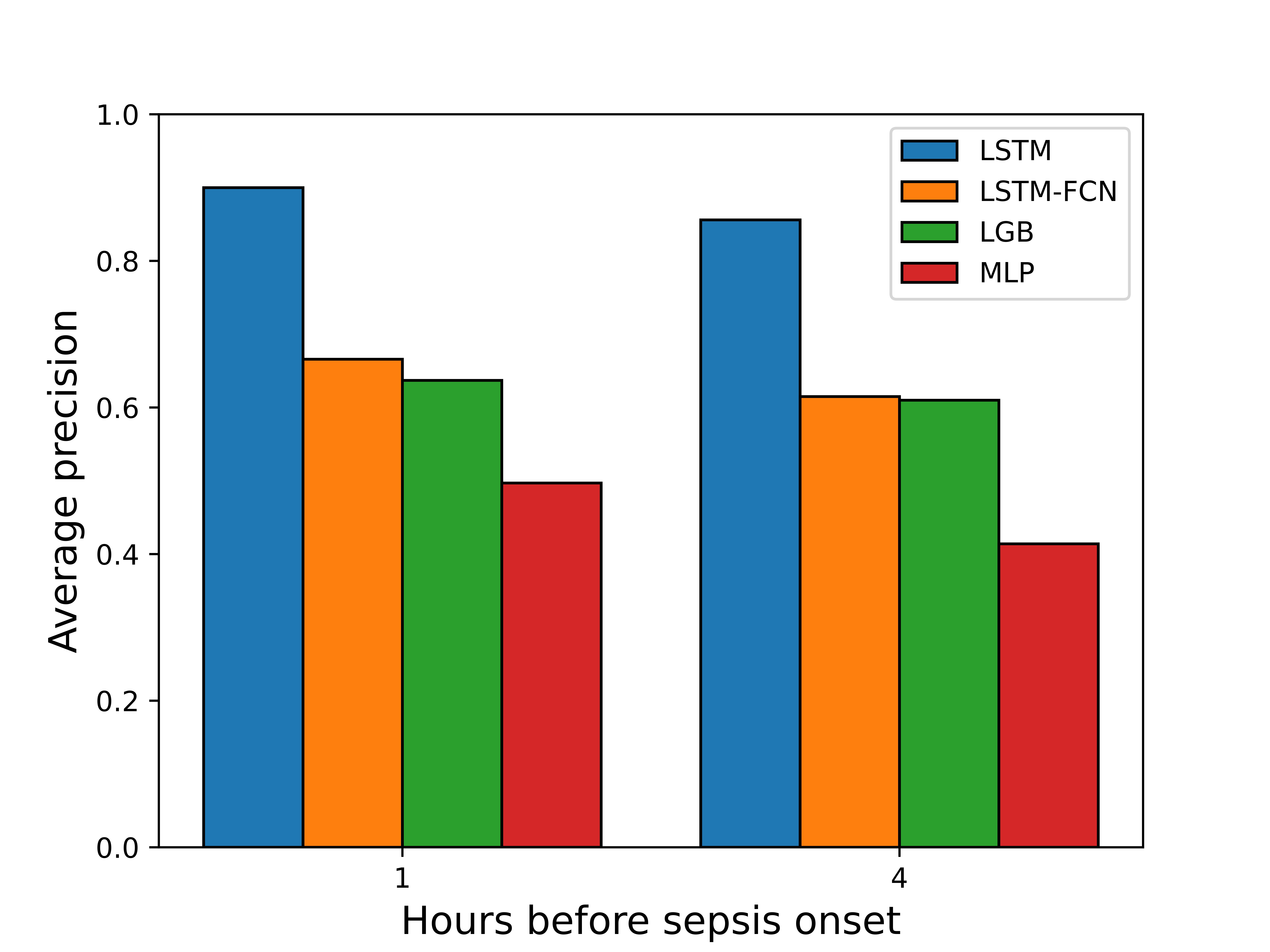}
         \caption{}
         \label{6b}
     \end{subfigure}
     \hfill
        \caption{Evaluation results of all models on the test dataset: (a) AUROC at different prediction windows; (b) AP at different prediction windows.}
        \label{fig 5}
\end{figure}

The LSTM network reached an AUROC of 0.96 and an mPA of 0.91 for evaluating a one-hour prediction window and an AUROC of 0.92 and an mPA of 0.90 for a four-hour prediction window (Figures \ref{fig 4}a and \ref{fig 4}b). The calibration curve shows the risk of sepsis development prediction using LSTM is mostly lower than the observed frequency of the onset. Although the LSTM model has the best performance among all other approaches, it has the largest architecture size (1541 kb). Also, its execution time is twice as much as the LSTM-FCN model. Therefore, the LSTM model performs the most computational complexity for the early prediction of sepsis.

\subsection{LSTM-FCN}

The most prominent point about the LSTM-FCN model is its acceptable performance, whereas a small architecture size (141 kb) as a deep learning model. In spite of this model size, it entails a long execution time of 0.16 ms, which can indicate the complexity of calculations. This model represented 0.824 and 0.800 AUROC, 0.666 and 0.615 AP for prediction windows of one- and four-hour, respectively.

\begin{table}[htbp]
    \centering
    \captionsetup{justification=centering}
    \caption{Performance metrics and specifications of the developed models.}
    \begin{tabular}{lcccccccc}
        \hline
        \textbf{Metrics} & \multicolumn{4}{c}{\textbf{One-hour}} & \multicolumn{4}{c}{\textbf{Four-hour}} \\
        & MLP & LGB & LSTM-FCN & LSTM & MLP & LGB & LSTM-FCN & LSTM \\
        \hline
        AUROC & 0.686 & 0.825 & 0.824 & 0.964 & 0.623 & 0.804 & 0.800 & 0.902 \\
        AUPR & 0.497 & 0.637 & 0.666 & 0.900 & 0.414 & 0.610 & 0.615 & 0.856 \\
        Sensitivity & 0.85 & 0.85 & 0.85 & 0.85 & 0.85 & 0.85 & 0.85 & 0.85 \\
        Specificity & 0.53 & 0.73 & 0.76 & 0.97 & 0.45 & 0.72 & 0.69 & 0.91 \\
        Accuracy & 0.73 & 0.74 & 0.75 & 0.93 & 0.71 & 0.72 & 0.72 & 0.88 \\
        Size (kb) & 202 & 196 & 141 & 1541 & 202 & 196 & 141 & 1541 \\
        Execution time (ms) & 0.09 & 0.004 & 0.16 & 0.32 & 0.09 & 0.004 & 0.16 & 0.32 \\
        \hline
    \end{tabular}
    \label{table 1}
\end{table}

\section{DISCUSSION}

Developing intelligence-based models in predicting sepsis onset have been previously studied. An extensive diversity of ML and deep learning models are proposed for in-hospital applications. However, the lack of these systems for out-of-hospital patients is evident. The reasons are manifold: first, there are no publicly available datasets of patients who are not admitted to an ICU or a ward, as it would be hard to collect data on different features from such patients. Second, most of the sepsis criteria are defined for in-hospital patients. More precisely, out-of-hospital criteria and scoring systems, such as qSOFA, are clarified that they should just be used for sepsis onset suspicions and prompt actions \cite{singer2016third}. Therefore, they cannot be replaced with in-hospital criteria, and their performance is not as accurate as in hospitals’ scoring systems. Subsequently, the labeling of such data would be a challenging task. Third, developing a reliable system to predict sepsis onset in advance based on available out-of-hospital features is a tough assignment.
On the other hand, the growing adoption of telehealth, as well as the accurate ability of wearable devices to measure signs, most importantly heart rate, and its computational capability, might lead to the development and implementation of intelligence systems for this task. Wearable devices could be a desirable means of monitoring people and surveillance patients by physicians as an appropriate and efficient device. More specifically, they potentially could serve as sepsis warning devices as well as their usual task. Besides, heart rate is noticed as one of the most influential, substantial, and common features for the detection of sepsis in both ML works and clinical scoring systems \cite{fleuren2020machine, moor2021early}. In this study, we have developed, evaluated, and analyzed several ML approaches for the prediction of sepsis using the heart rate feature. More deliberately, we have implemented four different methods for early sepsis detection: LGB, MLP, LSTM, and LSTM-FCN models. Having considered the restrictions of wearable devices, the GA has been employed in the first three models to fulfill the requirements by optimizing their architectures. Model performance, primary storage, and the complexity of calculation criteria are regarded for selecting the best methods parameters and comparing all models. We have considered the response time of each model in the face of a particular test data as a representation of the complexity of calculations. This criteria potentially comprises two factors: the number of calculations and complications of every computation inside the methods. Thus, the longer the execution time for data, the more complexity of calculations. Even though this statement is not necessarily true in any conditions, we can rely on it as there was a significant difference between the execution times of different models. In addition, the two mentioned factors have a direct correlation with battery usage. As a result, if the execution time takes longer, the required memory and battery usage in a wearable device will be higher, which is a crucial factor for such devices. These points have been considered in the GA algorithm to discover the best parameters for the models, which are reported in Table \ref{table 1}. As expected, these values are the same for both prediction windows since the structure of each model is similar in the different windows, and the nuance is in the weights of the models. The LSTM model shows the highest execution time with 0.32 ms, which means twice as long as the second-highest model (LSTM-FCN) and eighty times longer than the lowest model’s execution time (LGB). Furthermore, the MLP model has an execution time of 0.09 ms to make a prediction using twelve-hour heart rate sequence data. By contrast, LSTM has a superior accuracy among the models, with an accuracy of 93\% and 88\% for one- and four-hour prediction windows, respectively. The performance metrics of LSTM-FCN and LGB models are approximately similar, while MLP demonstrated not a promising performance in both windows. Of all metrics, specificity should be considered the most critical for sepsis prediction as the sepsis dataset is unbalanced, and automated intelligence decision models usually suffer low specificity (higher false alarm rate) in this task. The LSTM model presented an encouraging specificity in both prediction windows, and surprisingly, LGB had a higher specificity than the LSTM-FCN deep model in the four-hour window. It should be noted that the models’ performance decreased impressively when we tried to generalize them for longer prediction windows, such as six and twelve. Predictably, it stems from using just the heart rate feature for the prediction, which cannot preserve the efficiency in longer windows. 
Another considerable criterion for wearable device plug-ins is primary storage. Due to their restricted storage, we should ponder the size of developed models for real implementation. Similar to execution time, primary storage is also alike for both prediction time windows since the preservation of the models’ architecture. The LSTM model takes up a lot of storage by a large margin with 1541 kb. Following that, 202 kb and 196 kb have been taken by the MLP and LGB models. Interestingly, the LSTM-FCN approach, as a deep learning model, needs the lowest storage size (141 kb) for implementation.
Previous studies have shown that even a one-hour delay in treating sepsis is a consequence of an increment in mortality probability, emphasizing the importance of every hour in early prediction and medical intervention \cite{kumar2006duration, seymour2017time}. Unfortunately, most out-of-hospital cases come to the hospital after an incident of sepsis, which makes it more difficult for physicians to prevent the consequences. Studies revealed that sepsis patients who received prehospital medical care had lower odds of dying in the hospital \cite{rhee2017incidence}. Thus, potential patients would receive medical care and be hospitalized prior to the onset of sepsis using the developed ML models. Another promising application of the developed ML model could be to triage incident sepsis cases that come into the emergency department by analyzing patient data and identifying those in need of immediate attention. This approach could help to reduce delays in diagnosis and treatment, which is critical for improving patient outcomes. By detecting sepsis cases earlier, healthcare providers can initiate prompt investigations and provide appropriate interventions, such as antibiotics and fluid resuscitation, to improve outcomes and decrease the risk of complications.

\subsection{Limitation}
\subsubsection{Dataset, gold standard, and oversampling}

This study primarily utilizes the PhysioNet 2019 Challenge dataset, originally intended for ICU usage, for a non-ICU task. The use of this dataset presents inherent limitations due to its ICU-specific SOFA scoring system and lack of suitable out-of-hospital data, as finding sepsis cases outside the hospital setting is challenging. While the ICU-oriented data may enhance model precision, it restricts the generalizability of our findings to non-ICU environments. It would be beneficial for future studies to adapt the PhysioNet dataset with qSOFA labeling to assess model performance with out-of-hospital criteria and explore other machine learning models tailored for this context.
Additionally, the PhysioNet dataset, the only publicly available data for sepsis onset prediction, suffers from imbalances which we addressed through oversampling non-sepsis classes. This approach, however, complicates model development, especially when relying solely on limited heart rate data from wearable devices. Such data restrictions not only limit the diversity and volume of data but also constrain the range of features, primarily to heart rate signals, thus curtailing the predictive power of our models. 
Despite these challenges, there is significant potential for advancements as data collection technologies evolve and more comprehensive datasets become available. We anticipate that future research will leverage these developments to refine and expand our models to include a broader array of physiological parameters and patient demographics. By doing so, we can enhance the models' robustness and applicability, thereby extending the scope of machine learning applications in clinical sepsis prediction and potentially overcoming the current limitations.

\subsubsection{Bias and configuration}

As the proposed deep learning models operate in high-dimensional feature space, it is required to consider the related bias issues for any practical implementation. It has been shown that generalizing these kinds of models is not as easy as it seems at the development stage \cite{agniel2018biases,zech2018confounding}. Primarily, our method can be implemented in practice when this issue is carefully considered. That is, due to the variety of software and hardware configurations of wearable devices, the bias problem needs to be handled.

\section{CONCLUSION}

Despite sepsis’s high morbidity and mortality rate, intelligent, supportive systems with little variety were developed to help mitigate the consequences. In this study, we proposed ML-based methods for the prediction of sepsis onset one- and four-hour in advance. The purpose of the developed models, unlike the previous in-hospital supportive decision systems, is to be utilized outside of medical-related environments. Since nowadays wearable devices can detect heart rate accurately and reliably, we employed four distinct ML models and tried to optimize their structure using the genetic algorithm to predict sepsis based on a sequence of heart rate feature. In this regard, we considered the model’s performance, the complexity of calculations, and the model’s size as the criteria for developing the models. The promising results demonstrate that the developed LSTM model can predict sepsis onset in advance with fairly reliable performance using heart rate.


\bibliographystyle{unsrt}

\bibliography{sample}



\end{document}